%% file: main.tex
\documentclass{article}

\usepackage{arxiv}

\usepackage[utf8]{inputenc} 
\usepackage[T1]{fontenc}    
\usepackage{hyperref}       
\usepackage{url}            
\usepackage{booktabs}       
\usepackage{amsfonts}       
\usepackage{nicefrac}       
\usepackage{microtype}      
\usepackage{lipsum}		
\usepackage{graphicx}
\usepackage[numbers,sort&compress]{natbib}
\usepackage{doi}

\usepackage{amsmath}
\usepackage{tabularx}
\usepackage{placeins}
\usepackage{makecell}
\usepackage{longtable}
\usepackage{adjustbox}
\usepackage{threeparttable}

\usepackage{xcolor}         
\usepackage{multirow}

\newcolumntype{Y}{>{\raggedright\arraybackslash}X}

\usepackage{lineno}

\usepackage{subcaption} 
\usepackage[most]{tcolorbox}
\usepackage{listings}

\usepackage{changepage}
\usepackage{array}

\newcolumntype{C}{>{\centering\arraybackslash}X}

\lstdefinestyle{searchquery}{
  basicstyle=\ttfamily\footnotesize,
  breaklines=true,
  columns=fullflexible,
  keepspaces=true,
  showstringspaces=false
}

\newtcblisting{searchquerybox}[1]{
  breakable,
  colback=gray!4!white,
  colframe=gray!60!black,
  title={#1},
  listing only,
  listing options={style=searchquery}
}

\usepackage{pdfpages}
\usepackage{amssymb}   %
\usepackage{mathtools} %
\usepackage{amsthm}    %
\usepackage[capitalize,noabbrev]{cleveref} %
\usepackage[textsize=tiny]{todonotes} %
\usepackage{xspace}    %
\usepackage{colortbl}  %
\usepackage{algorithm} %
\usepackage{algorithmic} %
\usepackage{wrapfig}   %
\usepackage{sidecap}   %
\usepackage{soul}      %
\usepackage{enumitem}  %
\usepackage{multicol}  %
\usepackage{pifont}    %
\usepackage[normalem]{ulem} %
\usepackage{titletoc}  %

\newcommand{\ouralgo}{\textsc{SciLitBench}\xspace}

\definecolor{dodgerblue}{rgb}{0.12, 0.56, 1.0}

\title{\ouralgo: Benchmark and Design Principles for LLM-Powered Systematic Literature Reviews}

\author{Miguel Zabaleta$^1$, Baihan Lin$^{1,2,3,*}$  \\ \\
  $^1$ Department of AI and Human Health, Icahn School of Medicine at Mount Sinai, New York, NY, USA \\
  $^2$ Department of Psychiatry, Icahn School of Medicine at Mount Sinai, New York, NY, USA \\
  $^3$ Department of Neuroscience, Icahn School of Medicine at Mount Sinai, New York, NY, USA \\
    $^*$ Corresponding: \texttt{baihan.lin@mssm.edu} \\
}

\renewcommand{\headeright}{}
\renewcommand{\undertitle}{Manuscript }
\renewcommand{\shorttitle}{\ouralgo: Benchmark and Design Principles for LLM-Powered Systematic Literature Reviews}

\definecolor{paperlinkblue}{rgb}{0.01,0.31,0.65}
\hypersetup{
  colorlinks=true,
  linkcolor=paperlinkblue,
  citecolor=paperlinkblue,
  urlcolor=paperlinkblue,
pdftitle={\ouralgo: Benchmark and Design Principles for LLM-Powered Systematic Literature Reviews},
pdfsubject={q-bio.NC, q-bio.QM},
pdfauthor={Miguel Zabaleta},
pdfkeywords={systematic reviews, large language models, benchmark, literature review automation, data extraction},
}
\newcommand{\figref}[2][]{\hyperref[#2]{Figure~\ref*{#2}#1}}
\newcommand{\tabref}[1]{\hyperref[#1]{Table~\ref*{#1}}}
\newcommand{\suppfigref}[1]{\hyperref[#1]{Supplementary Figure~\ref*{#1}}}
\newcommand{\supptabref}[1]{\hyperref[#1]{Supplementary Table~\ref*{#1}}}

\begin{document}
\maketitle\begingroup\renewcommand{\thefootnote}{}\footnotetext{Data, code, and other artifacts are released at \url{https://github.com/linlab/SciLitBench}.}\endgroup

\begin{abstract}
Systematic reviews require sustained human judgment across thousands of records, yet existing evaluations of large language models (LLMs) typically examine review stages in isolation. We introduce SciLitBench, a multi-stage benchmark spanning title and abstract screening, full-text screening, and schema-guided data extraction, with 42,981 retrieved records, 1,012 full texts, and annotations for 888 included papers. Across 22 open-weight LLMs from six model families, explicit inclusion and exclusion criteria improve title and abstract screening $F_2$ by 28.8\%, while researcher-authored rationales improve full-text screening by 15\%. Data extraction reveals a different reliability regime: performance declines from 0.97 accuracy for publication year to 0.37 Jaccard overlap for computational approach, while the strongest models recover only 30\% of annotated evaluation evidence and 25\% of limitations. SciLitBench identifies a practical boundary between high-recall screening and evidence-complete extraction and provides a reproducible resource for evaluating LLM-assisted evidence synthesis.

\end{abstract}



Systematic literature reviews turn scattered studies into evidence that can guide science, medicine, and policy. Meta-analyses in public health and medicine, including studies of passive smoking and trans fats, have shaped guidance affecting millions of people \citep{cite3, cite4, cite5, cite6, cite7}. As the research literature grows, so does the need for such synthesis: an estimated 80 reviews are now published each day \citep{cite1, cite2}, yet producing each review remains slow and costly. A single high-quality review can take years and cost hundreds of thousands of dollars \citep{cite_cost}, because every stage requires consistent human judgment across thousands of records.

The workload of systematic reviews has made automation an active target for LLM-based scientific tools \citep{cite_aiindex, cite_llms1, cite_llms2, cite_llms3}. Prior studies have evaluated models for title and abstract screening, full-text screening, and data extraction \citep{cite_titleabstract1, cite_titleabstract2, cite_titleabstract3, cite_titleabstract4, cite_titleabstract5, cite_dataextraction1, cite_dataextraction2}, and related work has tested prompting strategies and ensemble methods \citep{cite_prompttuning1, cite_prompttuning2, cite_prompttuning3, cite_prompttuning4, cite_prompttuning5, cite_ensembles1}. Yet this evidence remains fragmented by review stage. Existing benchmarks isolate title and abstract screening, extraction from abstracts, or full-text effect inference, but none follows a single review from search through screening to extraction on a unified corpus \citep{synergy, deyoung2020evidence, nye2018pico, deyoung2021ms2}. \tabref{tab:related-benchmarks} compares these resources by review stage, document length, released rationales, and scale. This leaves three questions unresolved: \textit{Which design choices make review automation reliable? Does reliability transfer across title and abstract screening, full-text screening, and data extraction? And what changes when extraction targets shift from closed and unambiguous to open-ended evidence extraction?}

\begin{table}[tb]
  \begin{center}
        \setlength{\tabcolsep}{4pt}
        \renewcommand{\arraystretch}{1.05}
        \begin{tabular}{lccccr}
          \toprule
          Benchmark & Domain & Stages & Long & Rationale & Size \\
          \midrule
          \textbf{\ouralgo (ours)} & SR & TA+FT+DE & $\surd$ & $\surd$ & 43k TA; 1k FT; 17k DE \\
          SYNERGY \cite{synergy} & SR & TA & $\times$ & $\times$ & 169k recs; 26 SRs \\
          Evidence Inference \cite{deyoung2020evidence} & RCT & FT & $\surd$ & $\surd$ & 7k q--doc inst. \\
          EBM-NLP (PICO) \citep{nye2018pico} & RCT abs. & DE & $\times$ & span & 5k abs. \\
          MS$^{2}$ \citep{deyoung2021ms2} & Med SR & Summ. & $\surd$ & $\times$ & 470k docs; 20k sums \\
          BioASQ \citep{tsatsaronis2015bioasq} & Biomed & IR+QA & snip & $\surd$ & Yearly batches \\
          PubMedQA \citep{jin2019pubmedqa} & Biomed abs. & QA & $\times$ & $\times$ & 1k lbl; 211k auto \\
          \bottomrule
        \end{tabular}
  \end{center}
    \caption{\textbf{\ouralgo in context.} \ouralgo uniquely spans three SR stages (TA/FT/DE) on long PDFs with released inclusion rationales, enabling LLM-centric evaluation. Abbreviations: SR (systematic review), TA (title and abstract screening), FT (full-text screening), DE (structured data extraction), RCT (randomized controlled trial), IR (basic information retrieval), QA (question answering), Summ. (summarization).}
    \label{tab:related-benchmarks}
    \vspace{-2em}
\end{table}

These questions matter because the stages of a review do not test the same capability. Screening asks whether a record or full text should be included; extraction asks what evidence the paper contains, where the difficulty of that task depends on what the reviewer asks the model to recover. Prior extraction studies have often focused on low-ambiguity targets: clinical sample size, PICO elements, or effect direction \citep{de_ref1, de_ref2, de_ref3}. Those targets differ from what realistic review automation requires. The publication date of a paper is a piece of information that is easy to determine if it's right or wrong, but a paper's approach, reported results, or limitations can be expressed at different levels of granularity, and multiple phrasings may be correct. We call this range the ambiguity spectrum. At its open-ended end, correctness must be evaluated semantically rather than by exact matching, motivating an LLM-as-a-judge (LaaJ) protocol calibrated with human assessments \citep{zheng2023llmjudge,laaj_calibration}.

To address these gaps, we introduce \ouralgo, a multi-stage benchmark built from a real systematic review that we conducted on computational methods for literature-review automation. From 42,981 retrieved records, we curate a unified corpus spanning title and abstract screening, full-text screening, and schema-guided data extraction, with researcher-curated labels, inclusion rationales, and documented criteria at each stage. We evaluate 22 open-weight LLMs from six model families. We center the benchmark on open-weight models because scientific comparisons require systems whose evaluated state can be preserved and rerun. Hosted proprietary models may change in weights, inference infrastructure, safety layers, or serving behavior without those changes being fully observable to researchers, complicating longitudinal comparison and exact replication. Open-weight models are not free from training-data, implementation, or model-family biases, but their weights and inference configurations can be archived alongside prompts and outputs. This makes them particularly suitable for a benchmark intended to support reproducible comparison over time. To our knowledge, no prior study has evaluated this breadth of open-weight model families across all three stages of a single systematic-review pipeline at comparable scale.

\ouralgo is released as an open benchmark with frozen splits, versioned prompts and schemas, model outputs, scoring scripts, and calibration artifacts. This work makes five contributions:

\begin{enumerate}[topsep=1pt, itemsep=1pt, parsep=0pt, partopsep=0pt]
    \item A multi-stage, open benchmark for review automation (\ouralgo), with frozen splits and documented protocols spanning abstract screening, full-text decisions, and extraction. 
    \item Comprehensive evaluation of 22 open-weight LLMs from six model families with standardized prompts and metrics, revealing robust gains from explicit inclusion/exclusion prompting and researcher rationales, and showing how current open-weight LLMs fall short of reliable data extraction in ambiguous tasks.
    \item A data extraction benchmark for systematic reviews over 888 full texts, spanning six fields from canonical attributes (year, domain, \ldots) to open-ended evaluation results and limitations, revealing substantial headroom for improvement under real-world ambiguity.
    \item A fully reproducible schema-guided extraction and evaluation workflow, combining iteratively-developed annotation schemes with calibrated LaaJ scoring, released with the schemas, prompts, outputs, parsed JSONs, gold labels, and code for direct comparison and reuse.
    \item A researcher-in-the-loop harmonization workflow that maps open-ended extractions into analysis-ready variables for downstream analysis of the review-automation literature.
\end{enumerate}

Across these models, screening reliability improves when instructions make inclusion criteria, exclusion criteria, and researcher rationales explicit. Data extraction reveals a different reliability regime: models perform well on low-ambiguity fields such as publication year, but performance deteriorates as targets require schema-dependent categorization or recovery of open-ended scientific evidence. Together, these stages expose a practical boundary between tasks that can be made dependable through protocol design and tasks that still require substantial researcher verification. We release the open-ended annotations in both raw and harmonized forms, giving researchers a dataset of the review-automation literature that can be queried, studied, and used to generate new hypotheses.

\section{Results}

\subsection{\ouralgo connects screening and extraction in a single review pipeline}

\ouralgo is built from a real systematic review of computational methods for literature review automation (a review of how artificial intelligence and machine learning methods are used to automate parts of the review process itself). Rather than repurposing existing datasets or constructing synthetic evaluation tasks, we use this review as both a working corpus and a benchmark, evaluating LLMs on the same documents, decisions, and ambiguities that human reviewers actually face. Because the review studies review automation itself, the corpus spans computer science, clinical research, social science, and beyond, covering every stage of the review pipeline and every class of computational approach. This breadth lets us probe model behavior across disciplines and methodological traditions rather than within a single narrow domain.

\begin{figure}[tbp]
\centering
\includegraphics[
  width=0.8\textwidth
]{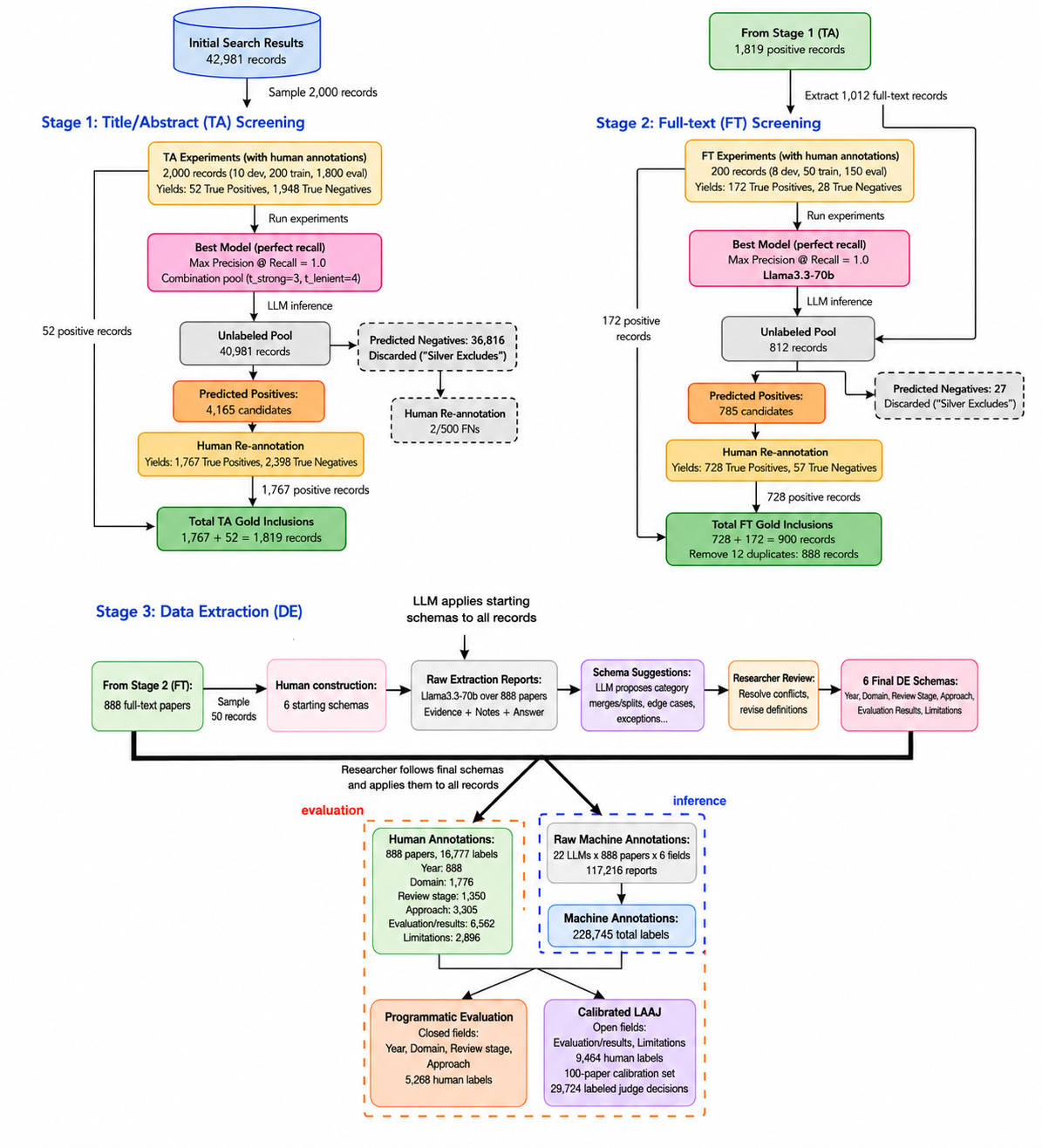}
\caption{\textbf{Construction of \ouralgo across title/abstract screening, full-text screening, and data extraction.}
\textbf{Stage 1: Title and abstract (TA) screening.} From 42,981 retrieved records, 2,000 were manually annotated and divided into development, few-shot, and evaluation subsets. Model and ensemble experiments were used to identify the highest-precision configuration achieving perfect recall on the evaluation split. This configuration was applied to the remaining 40,981 records, yielding 4,165 predicted positives for manual re-annotation and 36,816 predicted negatives retained as silver exclusions. Manual review of the predicted positives identified 1,767 additional inclusions and 2,398 exclusions; combined with the 52 inclusions from the initial annotated set, this produced 1,819 title/abstract inclusions. A random sample of 500 silver exclusions was additionally re-annotated to estimate false-negative contamination.
\textbf{Stage 2: Full-text (FT) screening.} Full text was retrieved for 1,012 records carried forward from Stage 1. A manually annotated set of 200 papers was used for full-text model evaluation and model selection. The highest-precision configuration achieving perfect recall was then applied to the remaining 812 papers, producing 785 predicted positives for manual re-annotation and 27 silver exclusions. Manual review identified 728 additional inclusions and 57 exclusions. Combined with the 172 inclusions from the initial annotated set and after removing 12 duplicates, this yielded 888 papers for data extraction.
\textbf{Stage 3: Data extraction (DE).} Schema development began from researcher-defined extraction goals and a 50-paper sample used to construct six initial schemas. Llama 3.3-70B generated raw extraction reports across the corpus, from which a collaborator pass proposed category merges, splits, edge cases, and exceptions. Researchers reviewed these suggestions, resolved conflicts, and finalized six extraction schemas: publication year, domain, review stage, approach, evaluation results, and limitations. Researchers then applied the finalized schemas to all 888 papers, yielding 16,777 human annotations. The same schemas were used to obtain machine annotations from 22 LLMs. Closed fields were evaluated programmatically, whereas the open-ended evaluation-results and limitations fields were assessed using a calibrated LLM-as-a-judge protocol based on a 100-paper calibration set and 29,724 labeled judge decisions. Solid arrows indicate the principal data flow through the review pipeline; dashed boxes denote evaluation or audit components rather than additional review stages.}
\label{fig:ta_ft_de_diagram}
\end{figure}

Starting from a librarian-optimized search across PubMed, Semantic Scholar, and Scopus, we retrieved 42,981 candidate records. The benchmark spans the three stages of the review pipeline (\figref{fig:ta_ft_de_diagram}): title and abstract screening, full-text screening, and data extraction. For each stage, we release evaluation splits, human-verified gold labels, and inclusion rationales. As part of the annotations release, title and abstract screening covers the full set of 42,981 records, with 1,819 confirmed inclusions; full-text screening covers 1,012 retrieved full texts, with 888 confirmed inclusions; data extraction covers all 888 included papers annotated under finalized schemas across six fields (year, domain, review stage, approach, evaluation results, and limitations), yielding 16,777 structured data elements that span an ambiguity spectrum from canonical attributes to open-ended interpretive descriptors. Studying the three stages together on a single corpus lets us examine how performance and errors propagate across the pipeline, a question that no prior benchmark has been positioned to answer.

The full schema construction and inference workflow is shown in \figref{fig:ta_ft_de_diagram}. We first manually annotated a sample of 2,000 records for title and abstract screening and 200 papers for full-text screening, using the resulting labels both for evaluation and for selecting the highest-precision configuration that achieved perfect recall. This configuration was then applied to the remaining unlabeled records, and predicted positives were manually re-annotated to produce additional gold inclusions, while predicted negatives are released as silver excludes. For data extraction, the schema itself was developed iteratively in collaboration with researchers, with extractor and collaborator models surfacing candidate categories that were then refined and standardized; final gold labels were produced under the finalized schemas. 

To establish label trustworthiness, we conducted two complementary audits and include an annotation manual documenting the stage-specific rules, edge cases, and examples used to make screening decisions at both stages. First, we re-annotated stratified random subsets of the gold splits 8–12 weeks after the initial labeling to assess intra-annotator consistency: on title and abstract screening we observed 99.6\% agreement (Cohen's $\kappa=0.992$) over 250 items, and on full-text screening we observed perfect agreement ($\kappa=1.00$) over 50 papers. Second, to assess potential contamination of the silver excludes with missed positives, we manually reviewed a random sample of 500 records from the 36,816 title and abstract silver excludes and found only 2 false negatives (0.4\%). These audits indicate high internal consistency in the re-annotated subsets and a low estimated false-negative rate in the sampled silver exclusions; they do not provide independent inter-annotator validation across all benchmark fields.

For screening stages, we report precision and recall together with $F_2$, as preserving relevant papers is crucial to conducting a systematic review. For data extraction, publication year is evaluated by accuracy, the three multi-label closed fields by Jaccard overlap, and the open-ended fields by calibrated semantic evaluation. Full evaluation details are provided in Methods. \tabref{tab:related-benchmarks} shows that \ouralgo is the only benchmark in this comparison spanning title and abstract screening, full-text screening, and data extraction, while also including long-document tasks and released rationales.

\subsection{Explicit decision protocols and calibrated voting improve title and abstract screening}

Title and abstract screening improves substantially when the review decision itself is encoded explicitly in the prompting and aggregation protocol. To quantify these effects, we evaluated 22 open-weight models across six families and parameter scales under three prompting variants, with and without few-shot demonstrations, and examined majority-vote ensembles over the strongest configurations. A single review may require human judgment on tens of thousands of records, making this stage a major source of reviewer workload. Results for every model and configuration are reported in \supptabref{tab:stage_A}; prompts are detailed in Supplementary Information, Section~\ref{appendix:prompts1}.

A clear scaling trend emerges: screening performance generally increases with model scale (\figref[a]{fig:scaling_ft_reliability}). Mixture-of-experts and reasoning-oriented variants can perform well even at moderate sizes, and the overall association between model capacity and reliability is positive. Model capacity is therefore an important contributor to screening performance, but it does not determine performance on its own. We also find that specific prompt designs reliably improve screening performance.

\begin{figure}[tbp]
\centering
\includegraphics[width=0.8\textwidth]{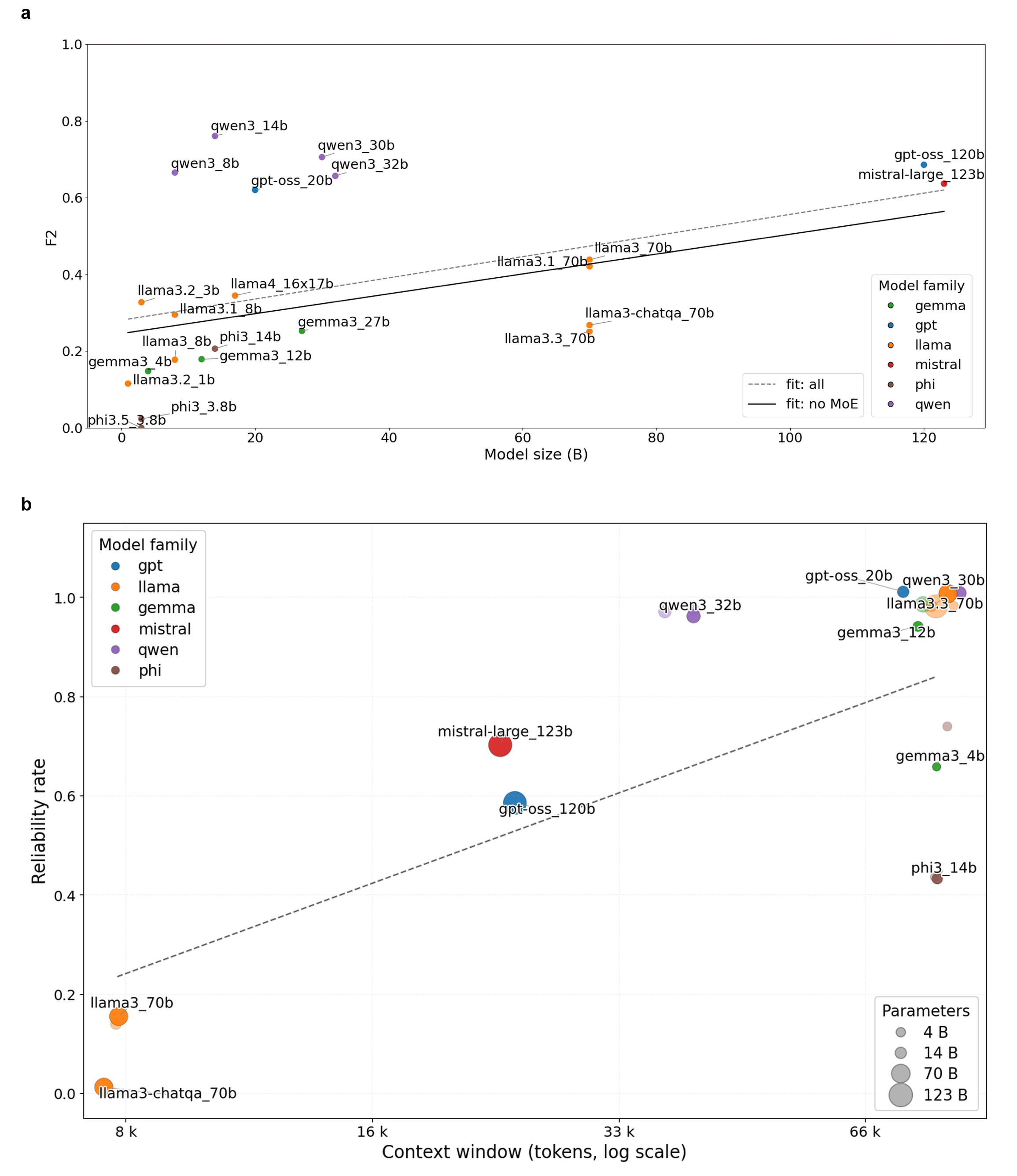}
\caption{\textbf{Model capacity and full-text reliability across screening stages.} \textbf{a)} Title and abstract screening performance scales with model size across 22 open-weight LLMs. Smaller models (\textless20B) generally perform poorly, while larger models achieve higher reliability; reasoning-oriented variants such as Qwen3 and gpt-oss reach strong performance despite moderate size. \textbf{b)} Full-text screening reliability depends jointly on parameter count and context window. Models with sufficient scale and long contexts achieve near-perfect reliability, while models with large context windows but insufficient parameters, or high capacity but limited context, often fail to produce parseable yes/no decisions.}
\label{fig:scaling_ft_reliability}
\end{figure}

We compared three prompt variants: a generic classification query, a prompt that explicitly asks whether the abstract meets inclusion criteria, and a prompt that additionally requires explicit exclusion reasoning. Inclusion prompting yields a 17.0\% relative gain in $F_2$; adding explicit exclusion reasoning raises this to 28.8\% (\figref[a]{fig:screening_protocol_design}). The pattern holds across model families and sizes, establishing prompt design as a simple but powerful lever for improving screening reliability: it requires no fine-tuning or additional compute, and any practitioner can apply it directly.

Few-shot demonstrations, by contrast, are less consistently helpful. Varying both the number of examples (2 or 5) and the proportion of positive cases (0.0, 0.5, 1.0), we find that few-shot performance does not exceed the zero-shot baseline on aggregate. At the model level, several strong configurations reach their peak $F_2$ with few-shot examples, suggesting that while examples are not a universally reliable lever, they can be helpful for particular models.

At the right threshold, majority voting improves on both the averaged baseline and the best single model. We defined two pools: a strong pool of the top five configurations ranked by $F_2$, and a lenient pool of the top ten configurations ranked by recall (subject to minimum precision of 0.2). Within each pool, we varied the vote threshold required for an inclusion decision. Both pools show an inverted-U pattern (\figref[c,d]{fig:screening_protocol_design}): at low thresholds nearly all candidates are included, while at high thresholds the ensemble becomes too restrictive. The balanced configurations (threshold 3 in the strong pool and threshold 9 in the lenient pool) achieve $F_2$ of 0.808 and 0.803 respectively, outperforming both their averaged baselines (0.766 and 0.646) and the best single models (0.786 and 0.749).  
For downstream curation of the unlabeled pool, we selected a combined configuration that triggers inclusion if either pool crosses its threshold (strong threshold 3, lenient threshold 4), achieving perfect recall (1.000) at precision 0.275, a deliberately conservative screening configuration that prioritizes capturing every relevant paper. While this setup reduces the workload of human review by approximately 90\% relative to fully manual screening, it also implies that human re-annotators must still examine a substantial number of false positives. We therefore also report the precision-recall tradeoff at less stringent recall thresholds. At recall \(\geq 0.98\), the best combined-pool configuration achieves precision 0.411 with 93.7\% workload reduction; at recall \(\geq 0.95\), it achieves precision 0.450 with 94.3\% workload reduction. Relaxing recall by 2–5 percentage points therefore raises precision from 0.275 to over 0.40, substantially reducing the number of irrelevant abstracts requiring manual inspection while preserving nearly all relevant literature. The choice between these recall targets is itself a methodological decision that practitioners can make based on their tolerance for missed studies.

\begin{figure}[tbp]
\centering
\makebox[\textwidth][c]{\includegraphics[width=1.10\textwidth]{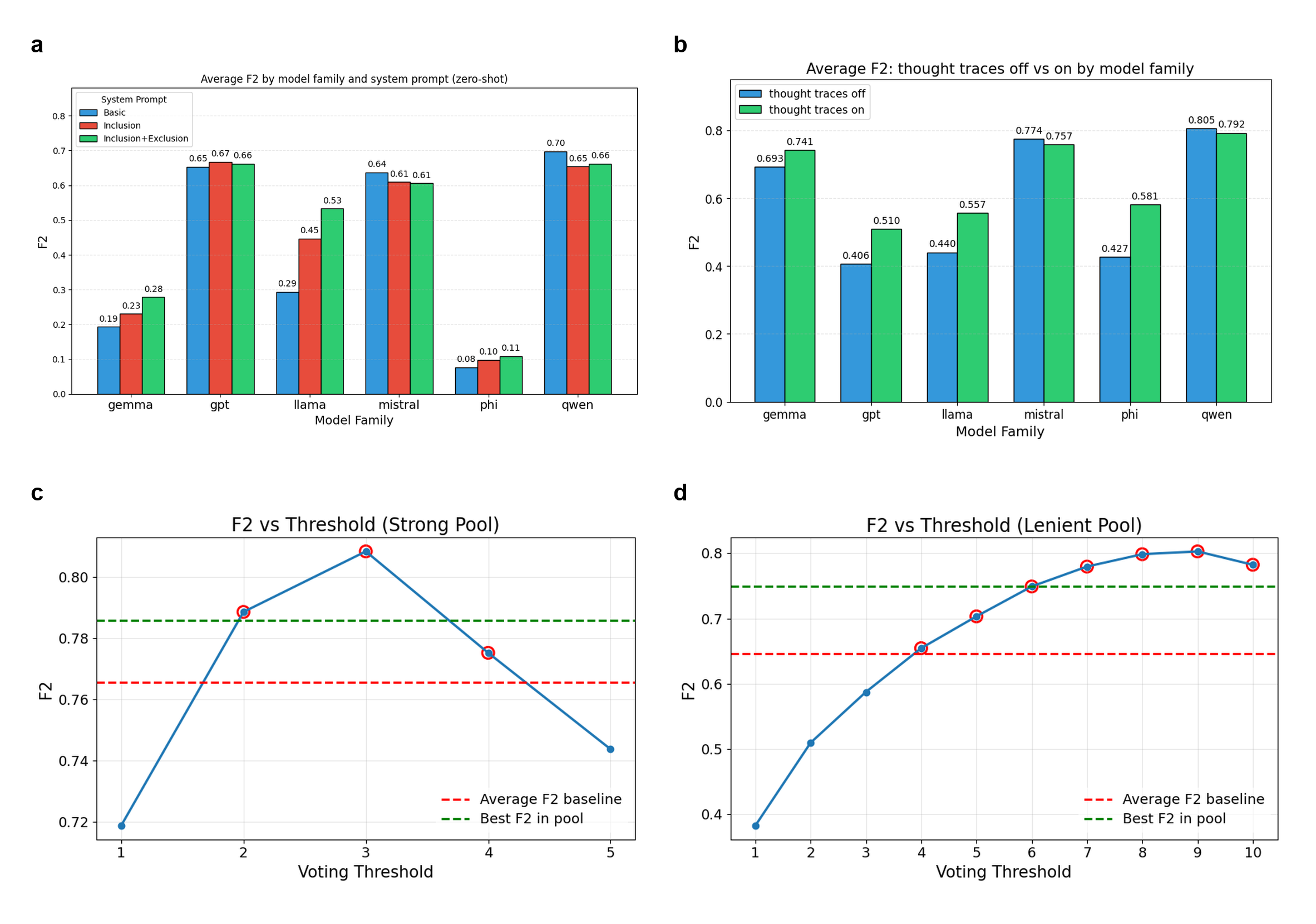}}
\caption{\textbf{Protocol choices that improve screening reliability.} \textbf{a)} In title and abstract screening, prompts that make the inclusion criteria explicit improve average $F_2$ by 17.0\% over a generic classification prompt; adding exclusion reasoning raises the total gain to 28.8\%. \textbf{b)} In full-text screening, adding researcher rationales to few-shot examples raises average $F_2$ from 0.551 to 0.633, a 15\% relative improvement. \textbf{c) and d)} Majority voting improves title and abstract screening only at suitable thresholds. In the strong pool, threshold 3 maximizes $F_2$ at 0.808; in the lenient pool, threshold 9 maximizes $F_2$ at 0.803, exceeding both the average baselines and the best single models.}
\label{fig:screening_protocol_design}
\end{figure}

\subsection{Long-document reliability depends on model capacity and context, while research rationales improve full-text screening}

Full-text screening remains sensitive to long-context reliability, but researcher-authored rationales provide a strong and reusable alignment signal. Inputs are long-document PDFs rather than short abstracts; inclusion criteria are more nuanced and often require integrating evidence from methods, results, and discussion sections; and stricter format adherence is needed if the model's outputs are to be reliably consumed downstream. We evaluated the same 22 models under these conditions, examining how capacity and context length jointly determine reliability, whether structured reasoning improves alignment with inclusion criteria, and whether researcher-authored rationales (verbatim evidence quotes paired with one-sentence justifications) can guide models toward more consistent decisions. Results for every model and configuration are reported in \supptabref{tab:stage_B}; prompt wording is detailed in Supplementary Information, Section~\ref{appendix:prompts2}.

A central finding is that output reliability (whether the model adheres to the instruction to answer with a strict yes/no decision) is itself a primary failure mode at this stage, jointly determined by parameter count and context window (\figref[b]{fig:scaling_ft_reliability}). Larger models with long context windows approach perfect reliability, whereas smaller models or those with shorter context windows can fall to rates near 0.4. We therefore adopt a two-step prompting scheme throughout this stage, in which an auxiliary refinement model converts free-form responses into strict yes/no outputs. This lets us evaluate screening decisions separately from parseability failures.

A chain-of-thought variant that requires the model to reason through the inclusion criteria before deciding yields a 3.8\% relative gain in $F_2$ (0.681 to 0.708) over an unstructured prompt. The gain is modest but consistent, indicating that structured reasoning can sharpen alignment with task instructions even in constrained binary settings (\suppfigref{fig:ft_cot}).

A substantially larger gain comes from incorporating researcher-authored rationales into few-shot demonstrations. We refer to these as researcher rationales, or ``thought traces'', because they pair the inclusion decision with quoted evidence and a concise justification; they should not be confused with model-generated chain-of-thoughts. Adding researcher rationales to few-shot examples raises average $F_2$ from 0.551 to 0.633, a 15\% relative improvement (\figref[b]{fig:screening_protocol_design}). The pattern holds across model families and is consistent with a broader design principle: when inclusion criteria are nuanced, examples are most useful when they include the reasoning that produced the decision, not just the decision itself. Researcher rationales can therefore serve both as review documentation and as an alignment signal for model-assisted screening, although producing them still requires reviewer effort.

The best single configuration, Llama 3.3-70B in zero-shot with the unstructured prompt, achieves $F_2$ of 0.981 at perfect recall (precision 0.910). We use this configuration to score the remaining unlabeled full texts and manually re-annotate predicted positives for downstream extraction corpus curation, mirroring the workflow used for title and abstract screening.

No ensembling configuration we tested (neither majority voting over strong or lenient pools, nor token-length-aware routing) improved over this single-model baseline on our corpus. To assess whether this reflects a general limitation of ensembling at this stage or a ceiling effect specific to our corpus, we ran an analysis through resampling: across 300 random pool configurations of 2–10 models with 70/30 train/validation splits, we evaluated the best single model, best majority-vote threshold, and best token-aware routing configuration. Under these more typical performance regimes, token-aware routing achieved a mean $F_2$ of 0.968 (95\% confidence interval 0.967–0.970), substantially and significantly outperforming both majority voting (0.926, CI 0.911–0.939; paired repeat-level test, $p < 10^{-8}$) and the best single model (0.877, CI 0.861–0.890). The negative ensemble result on our corpus therefore reflects an unusually strong single-model baseline rather than a general property of ensembling. Token-aware routing remains a useful design choice when single-model performance is more typical.

\subsection{Open-ended targets expose the limits of data extraction}

Data extraction marks the principal reliability boundary in the review pipeline: performance remains high for fixed bibliographic attributes but deteriorates sharply as extraction requires researcher-defined categorization and open-ended evidence recovery. Some targets have a single, easily checked answer, such as publication year. Others ask for higher-level descriptions of a paper: what was automated, how it was automated, what worked, and what failed. These targets are harder because papers can describe similar ideas in different language, category boundaries can overlap, and correctness often depends on semantic equivalence rather than string matching. \ouralgo's six DE fields span this ambiguity spectrum, from year to domain, review stage, approach, evaluation results, and limitations. This section tests how far schema-guided extraction remains reliable as the target moves from a bibliographic attribute to open-ended scientific evidence.

\begin{figure}[tbp]
\centering


\includegraphics[width=\linewidth]
{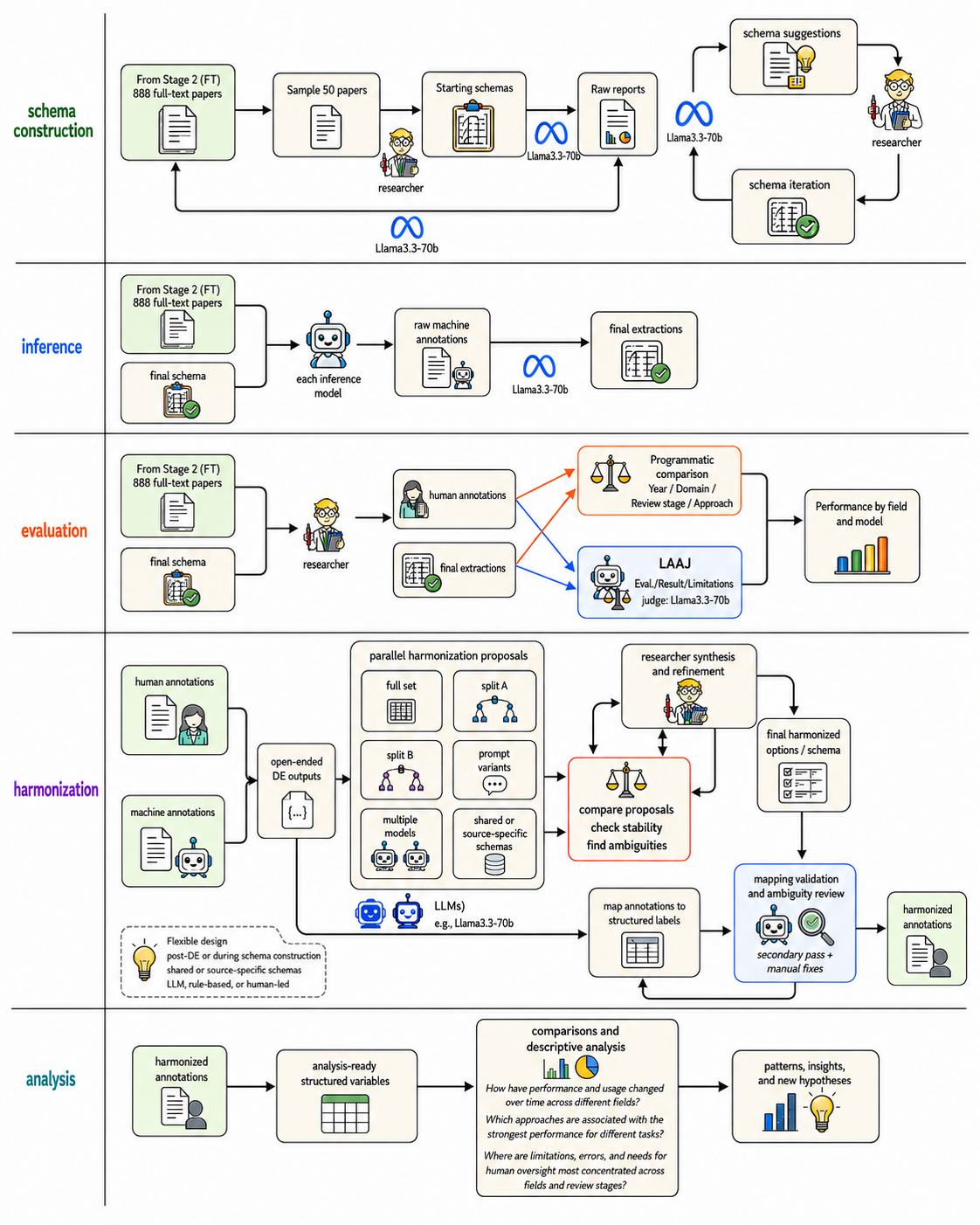}

\caption{\textbf{Data extraction and harmonization.} Data extraction begins with schema construction, applies each inference model to the included full-text corpus, and evaluates closed-label fields programmatically and open-ended fields with calibrated LaaJ evaluation. Harmonization is a downstream release step: raw free-text human and machine annotations are mapped into analysis-ready variables, giving researchers a rich dataset of the review-automation literature that can be queried and studied beyond model benchmarking.}
\label{fig:de_harmonization_workflow}
\end{figure}

For the closed-label fields, we use accuracy for year and Jaccard overlap for the three multi-label fields (Supplementary Table~
\ref{tab:de_closed}). The strongest open-weight models reach 0.97 on year. Scores are lower for domain (0.70), review stage (0.61), and approach (0.37), where the model must follow schema annotations that deal with moderately ambiguous questions.

Open-ended fields require a different evaluation approach because exact matching would penalize correct paraphrases. We therefore use a calibrated large language model-as-a-judge protocol to score model extractions against gold annotations, with calibration on a labeled subset to adjust for judge bias \citep{laaj_calibration}. Calibrated recall measures how much gold evidence the model recovers; calibrated precision measures how much of the model's extracted evidence is supported after calibration. We report only supported estimates in the main text, covering 78\% of model-by-field-by-metric combinations, with the full support criteria and unsupported entries in Methods and Supplementary Information.

For evaluation results and limitations, evidence is heterogeneous, paraphrased, and often difficult to compare by string matching. Across the evaluated open-weight models, the highest supported calibrated recall is 0.30 for evaluation results and 0.25 for limitations, meaning that no evaluated configuration recovers more than approximately one third of the annotated evidence under semantic matching (Supplementary Table~\ref{tab:de_open}). Some configurations achieve substantially higher precision, with field-wise maxima of 0.92 for evaluation results and 0.82 for limitations, but these maxima occur in selective configurations that recover relatively little evidence. Thus, high precision and broad evidence coverage do not coincide in the evaluated systems. The resulting failure mode is primarily one of incomplete evidence recovery rather than an inability to produce plausible structured outputs.


Finally, to examine how stage-level errors compound, we ran a cascade experiment with Llama 3.3-70B on 1,000 papers. Title and abstract screening recovered 97.6\% of relevant records. After subsequent screening, 33 papers reached data extraction; only 4 of these 33 papers were fully correct across the evaluated extraction fields, corresponding to 12.1\% complete-extraction accuracy conditional on reaching the extraction stage. The cascade therefore preserves most relevant records during early screening but loses substantial completeness when the task shifts from inclusion decisions to structured and interpretive evidence recovery.


\section{Discussion}

\ouralgo identifies a practical reliability boundary in LLM-assisted systematic review: screening can be made dependable through protocol design and recall-oriented model selection, whereas evidence-complete extraction remains substantially less reliable as targets become more interpretive. This distinction matters because the two stages fail in different ways. Screening primarily requires preserving relevant records under asymmetric error costs; extraction requires deciding what information counts as evidence, how that evidence should be represented, and whether differently worded statements are semantically equivalent. Treating both as generic applications of LLM capability obscures the design choices that determine whether automation is scientifically useful.

\textbf{Protocol design can matter as much as model scale.} Larger models generally performed better in title and abstract screening, and both capacity and available context shaped output reliability for full texts. Yet some of the largest practical improvements came from how the review decision was represented rather than from model scale alone. Explicit inclusion and exclusion criteria improved title and abstract screening, researcher-authored rationales strengthened full-text demonstrations, and calibrated voting allowed the operating point to be shifted toward the high recall required in evidence synthesis. These findings suggest that model selection should not be separated from decision-protocol design: the relevant unit of evaluation is the review system, including instructions, evidence representation, aggregation rule, and human verification policy, rather than the language model alone.

\textbf{Reliable screening is an asymmetric decision problem rather than a conventional classification task.} A false positive increases reviewer workload, whereas a false negative can remove a relevant study from every downstream analysis. The appropriate operating point therefore depends on the review team's tolerance for missed evidence. In our title and abstract experiments, perfect recall was achievable at lower precision, whereas small reductions in recall substantially increased precision and workload reduction. There is consequently no single universally optimal threshold. A useful review system should expose this tradeoff explicitly and allow researchers to choose the operating point appropriate to the scientific consequences of omission.

\textbf{Open-ended extraction exposes a different failure mode: incomplete evidence coverage.} Publication year can usually be represented by a single canonical value, whereas approaches, evaluation findings, and limitations depend increasingly on researcher-defined categories, granularity, and semantic interpretation. Performance deteriorated along this progression, and for the two open-ended fields no evaluated model recovered more than approximately one third of the annotated evidence. Importantly, some models remained precise by extracting only a small subset of what was present. Fluency or local factual correctness is therefore insufficient for evidence synthesis: a system may produce a convincing structured summary while silently omitting much of the evidence relevant to the review question.


\textbf{Human judgment moves upstream rather than disappearing.} The benchmark does not treat automation as the removal of researchers from the review process. Researchers define inclusion criteria, determine the useful granularity of extraction, refine schemas, document rationales, and adjudicate whether extracted statements are supported by the source literature. These decisions are especially consequential for open-ended fields, where multiple representations can be defensible. The harmonization layer makes this dependence explicit by preserving raw evidence while separately mapping it into analysis-ready variables. A productive role for LLMs may therefore be to reduce repetitive screening and extraction work while keeping epistemic decisions inspectable and researcher-controlled.


\textbf{Reproducibility is a design requirement for scientific benchmarks.} We center the main evaluation on open-weight models because their evaluated state can be preserved alongside prompts, inference settings, and outputs. This does not make open-weight systems intrinsically unbiased or more scientifically valid, and differences in training data, implementations, quantization, and model families remain important sources of variation. It does, however, permit future researchers to rerun the same systems and distinguish changes in benchmarking methodology from changes in the underlying model. Proprietary systems can provide useful contemporaneous reference points, but evolving hosted implementations make them less suitable as the sole foundation for longitudinal benchmark comparison.

\textbf{The benchmark is deliberately broad within one review, but it does not establish universal performance across systematic reviews.} \ouralgo is derived from a single review of computational methods for literature-review automation. Because this meta-topic spans computer science, medicine, social science, and multiple review technologies, it provides substantial diversity at the document and methodological levels; however, all papers are ultimately judged against one review protocol. The observed design principles should therefore be interpreted as findings within this corpus rather than universal laws of review automation. In addition, our screening-label audits measure temporal intra-annotator consistency and sampled contamination of silver exclusions rather than independent inter-annotator validity across every field. The extraction schemas were developed through a researcher-in-the-loop process in which model-generated proposals helped surface candidate categories before final researcher decisions, so the resulting ontology necessarily reflects both the review question and that development process. Finally, open-ended evaluation relies on calibrated LaaJ judgments and is therefore limited by the quality and support of the calibration procedure. These constraints are documented in the released annotations, schemas, prompts, calibration artifacts, and raw outputs so that future work can test alternative definitions and evaluation procedures.

Together, these results suggest that progress toward automated evidence synthesis should not be measured only by whether a model can generate a plausible review output. The more consequential question is whether the system preserves relevant studies, recovers the evidence required by the research question, exposes uncertainty and omissions, and remains reproducible enough to be audited. \ouralgo provides a common substrate for studying these properties across review stages. We expect future systems to combine stronger models with retrieval, selective prediction, uncertainty estimation, and researcher verification, but the central requirement remains the same: automation should reduce review burden without making the evidentiary path less visible.

\section{Methods}

\subsection{Search strategy}

\ouralgo was built from a systematic review of computational methods for literature review automation. The search strategy was developed with librarian input and run across PubMed, Semantic Scholar, and Scopus. The query was intentionally broad to minimize the risk of missing relevant records. After database export and deduplication, the search yielded 42,981 candidate records. The full query, database-specific syntax, and search documentation are provided in Supplementary Information.

The review protocol specified which papers counted as relevant and how the criteria applied at title and abstract screening, full-text screening, and data extraction. At the screening stages, each record or paper was labeled for inclusion or exclusion. Included full-text examples could also be paired with short rationales, consisting of quoted evidence and a brief explanation of why the paper satisfied the inclusion criteria. We later used these rationales as "thought traces" in prompting experiments.

\subsection{Experimental settings}

The main evaluation used 22 open-weight LLMs spanning multiple families and capacity ranges, including Llama \citep{touvron2023llama}, Gemma \citep{team2024gemma}, Qwen \citep{bai2023qwen}, Mistral \citep{jiang2024mistral}, Phi \citep{abdin2024phi}, and gpt-oss \citep{agarwal2025gpt}. We evaluated each model under standardized zero-shot and few-shot prompting conditions, with prompt variants defined separately for each review stage. Prompt versions were frozen before evaluation and are released with the benchmark. Open-weight models were run locally through Ollama using the model tags reported in Supplementary Information.

Unless stated otherwise, inference used temperature $=0$, top-$p=0.9$, and maximum token limits large enough to avoid truncating required outputs. For long full-text inputs, we passed the full extracted text up to the model's context limit. To standardize full-text screening runtime and avoid hardware-dependent slowdowns, models with native context windows of at least 128k tokens were evaluated with an 80k-token cap. Mistral Large 123B and gpt-oss 120B were capped at 24k tokens so that all layers remained on GPU on the H100-80GB setup. The longest full-text prompt in the corpus was shorter than 80k tokens, so the 80k cap did not truncate any input.

For the frontier comparison in data extraction, we evaluated Claude Sonnet 4.6, Gemini 3.1 Pro Preview, and OpenAI GPT-5.4 on the four closed fields across all 888 papers, using the same finalized schemas and scoring protocol as for the open-weight models.

For title and abstract screening and full-text screening, we report precision, recall, and recall-skewed $F_2$:
\[
\mathrm{Precision}=\frac{\mathrm{TP}}{\mathrm{TP}+\mathrm{FP}}, \quad
\mathrm{Recall}=\frac{\mathrm{TP}}{\mathrm{TP}+\mathrm{FN}}, \quad
F_2=\frac{5\cdot \mathrm{Precision}\cdot \mathrm{Recall}}{4\cdot \mathrm{Precision}+\mathrm{Recall}}.
\]
We use $F_2$ as the primary model-selection metric because false negatives are more costly than false positives in systematic-review screening: a false positive adds human workload, while a false negative can remove a relevant study from all downstream stages. We also report workload reduction, defined as the fraction of records that a reviewer would not need to manually inspect after model filtering at a chosen screening threshold.

For screening model selection, we first identified configurations that achieved perfect recall on the evaluation split. Among those configurations, we selected the one with the highest precision. For full-text screening, we also report reliability rate, defined as the fraction of outputs that followed the required strict yes/no response format. Stage-specific model-selection procedures, voting pools, and resampling analyses are described in the corresponding screening subsections.

Data extraction metrics depended on field type. Publication year was scored as accuracy. Domain, review stage, and approach can contain multiple labels, so we used Jaccard overlap between predicted and gold label sets. Evaluation results and limitations were scored with calibrated recall and calibrated precision, which estimate recovered gold evidence and supported model-extracted evidence under the LaaJ protocol described below.

\subsection{Title and abstract screening}

For title and abstract screening, we manually annotated an initial seed set of 2,000 records. This stage of the workflow is shown in \figref{fig:ta_ft_de_diagram}. The seed set contained 52 included records and 1,948 excluded records, and was divided into development, few-shot examples, and evaluation subsets of 10, 200, and 1,800 records. The frozen split is released with stable identifiers so that future systems can be evaluated on the same records.

The title and abstract corpus was expanded through a staged human-AI workflow. We used the manually annotated sample to evaluate individual model configurations, few-shot variants, and majority-vote ensembles. For majority voting, we defined a strong pool comprising the five configurations with the highest $F_2$ and a lenient pool comprising the ten configurations with the highest recall among those with precision of at least 0.2. We then varied the number of votes required for inclusion and selected the highest-precision configuration that achieved perfect recall on the evaluation split.

For downstream curation, this selected configuration was the combination voting pool with three strong-model votes and four lenient-model votes ($t_{\mathrm{strong}}=3$, $t_{\mathrm{lenient}}=4$). It was applied to the 40,981 unlabeled records. Predicted positives were manually re-annotated, yielding 1,767 additional included records and 2,398 excluded records. Combining these with the 52 positives from the initial annotation set produced 1,819 title and abstract inclusions. Records predicted as negative are released as silver excludes. The resulting workload reduction was computed as the fraction of unlabeled records not manually re-annotated, \((40{,}981 - 4{,}165) / 40{,}981 = 89.8\%\), reported in the main text as approximately 90\%.

We audited title and abstract labels by re-annotating a stratified random subset after an 8 to 12 week interval. Re-annotation of 250 records yielded 99.6\% agreement and Cohen's $\kappa=0.992$. To estimate contamination in the title and abstract silver-exclude set, we manually reviewed 500 records sampled from the 36,816 records predicted as negative by the selected high-recall system. This audit found 2 false negatives.

\subsection{Full-text screening}

For the full-text screening experiments, we manually annotated an initial seed set of 200 papers. This stage follows the title and abstract workflow shown in \figref{fig:ta_ft_de_diagram}. The set contained 172 included papers and 28 excluded papers and was divided into development, few-shot examples, and evaluation subsets of 8, 50, and 150 papers. The frozen split is released with stable identifiers so that future systems can be evaluated on the same records.

Full-text screening followed the same expansion logic after full-text retrieval and parsing. From the title and abstract inclusions, we extracted or retrieved 1,012 full texts. We used the full-text annotation sample to select the highest-precision configuration with perfect recall, applied this model to the remaining unlabeled full-text pool, and manually re-annotated predicted positives. This process yielded 728 additional full-text inclusions, which were combined with 172 positives from the initial full-text annotation set. After removing 12 duplicates, the final full-text inclusion set contained 888 papers for data extraction.

We audited full-text labels by re-annotating a stratified random subset after an 8 to 12 week interval. Re-annotation of 50 papers yielded perfect agreement and Cohen's $\kappa=1.00$.

Some full-text outputs contained useful reasoning but did not conform to the required yes/no format, so we used a two-step prompting scheme in which an auxiliary refinement model converted free-form outputs into strict decisions. The full-text evaluation also included a resampling analysis to test whether ensemble methods could help under less saturated model-pool conditions. The selected single model already provided a strong recall-constrained screening rule for this corpus, leaving little room for improvement in the main expansion run; the resampling analysis therefore asked how voting and routing behaved across more varied pools of available models.

For this analysis, we sampled random model pools of 2 to 10 models, split papers 70/30 into train and validation sets, selected the best single model, best majority-vote threshold, and best token-aware routing configuration on the train split, and evaluated all three on validation. We repeated this procedure 300 times. Majority voting treats each model decision as an unweighted vote and applies a fixed threshold. Token-aware routing extends this idea by allowing the number of model outputs used for a decision to vary with document length and by fitting combination weights on the train split. The paired repeat-level test reported in Results compares these strategies across the repeated pool configurations.

\subsection{Data extraction}

Data extraction was performed over the 888 included full-text papers. The extraction workflow is shown in \figref{fig:de_harmonization_workflow}. We defined six fields to capture what was automated, how it was automated, what worked, and what failed: publication year, domain, review stage, approach, evaluation results, and limitations. The resulting fields span the ambiguity spectrum introduced in the main text. Year is a low-ambiguity bibliographic attribute. Domain is constrained by a predefined taxonomy. Review stage and approach require schema development because useful categories depend on the observed corpus. Evaluation results and limitations are open-ended fields whose correct annotations can differ in wording, granularity, and emphasis.

Schema development was therefore an iterative part of the review protocol. It began with researcher-defined extraction goals and a 50-paper sample used to draft starting schemas and field definitions. We then used Llama 3.3-70B throughout the data extraction workflow as the fixed extraction and collaborator model, chosen because it was the strongest single model across our experiments. The model generated raw extraction reports over the corpus, including candidate evidence, notes, and answer fields. A collaborator pass proposed category merges, splits, edge cases, and exceptions from these reports. The researcher reviewed these suggestions, resolved conflicts, revised definitions, and produced the final schemas. Models could surface candidate structure, but the final annotation schemes reflected researcher choices about extraction scope, granularity, and category boundaries.

The final schemas and annotation manuals are released with field definitions, expected output formats, and edge-case guidance. Domain annotations use the ASJC taxonomy \citep{asjc_codes}. Review stage and approach use schemas developed from the corpus distribution. Evaluation results and limitations retain free-text evidence items under schema-defined categories, allowing semantic equivalence to be evaluated without forcing all evidence into a small controlled vocabulary.

For each included full text and each inference model, the data extraction prompt requested a structured report under the finalized schema. The report contained evidence from the paper, notes, and a final answer. Model outputs were parsed into JSON and validated against the expected field structure. Raw reports, parsed outputs, closed-field parsing and scoring scripts, prompts, schemas, LaaJ judge outputs, and calibration artifacts are released so that scoring decisions can be inspected and rerun.

Invalid or unparsable outputs were treated as errors for the affected fields. For closed-label fields, an invalid field received no credit. For open-ended fields, invalid or missing items could not be matched to gold evidence and therefore reduced recall.

For the multi-label closed fields, Jaccard overlap was computed between the predicted label set $\hat{Y}$ and the gold label set $Y$:
\[
J(\hat{Y},Y)=\frac{|\hat{Y}\cap Y|}{|\hat{Y}\cup Y|}.
\]

Evaluation results and limitations were evaluated with a calibrated large language model-as-a-judge protocol because exact string matching would penalize correct paraphrases. The judge compared model-extracted items with gold items within the same schema category, assessed semantic equivalence, and checked whether each item was supported by the source paper. Gold items with no supported match counted as missed evidence. Model items with no gold match were judged separately to determine whether they were still correct, supported extractions.

For the open-ended fields, calibrated recall and calibrated precision were computed as:
\[
\mathrm{Recall}_{\mathrm{cal}}=\frac{\mathrm{matched\ gold}}{\mathrm{gold\ items}}, \quad
\mathrm{Precision}_{\mathrm{cal}}=\frac{\mathrm{matched\ machine}+\mathrm{correct\ unmatched}}{\mathrm{machine\ items}}.
\]
Calibration follows \citet{laaj_calibration}. Judge outputs are calibrated on a labeled subset to adjust for judge sensitivity and specificity, then applied to model outputs for the full evaluation set. We calibrated the judge using a 100-paper subset, consisting of 29,724 labeled judge decisions across models. The LaaJ prompts, calibration labels, calibration computations, and aggregate support-status tables are released with the benchmark.

We classified each model-field-metric estimate as supported, unstable, or unsupported according to calibration quality. Let $q_0$ and $q_1$ denote the estimated judge specificity and sensitivity, and let $m_0$ and $m_1$ denote the numbers of calibration negatives and positives. An estimate was supported when both calibration classes were observed ($m_0>0$ and $m_1>0$), the adjusted judge signal exceeded 0.10 ($q_0+q_1-1>0.10$), and the analytic confidence-interval width was below 0.50.

Estimates were marked unstable when the denominator was positive but the calibration signal or interval width failed this support threshold. Estimates were unsupported when the denominator was zero or the calibration quantities could not be estimated. In total, 69 of 88 model-field-metric combinations are supported, 9 are unstable, and 10 are unsupported. Main-text tables report supported estimates and mark unsupported entries; full criteria and all unsupported or unstable estimates are provided in Supplementary Information.

\subsection{Benchmark release and harmonization}

\ouralgo is released as a reproducible benchmark package. The release includes frozen splits, stable identifiers, gold labels, silver labels, inclusion rationales, prompts, schemas, model outputs, parsed JSON outputs, scoring scripts, LaaJ prompts, and calibration artifacts. It also includes scripts for reproducing evaluation tables and figure inputs where redistribution is permitted.

For the open-ended fields, the benchmark evaluation uses free-text annotations, with calibrated LaaJ providing semantic comparison between gold and model outputs. These raw annotations preserve the original evidence statements and are released for users who want to inspect or rerun the evaluation. However, they are less convenient for researchers who want to aggregate the review-automation literature itself, because similar findings can be expressed with different wording and granularity. We also release a harmonized layer that maps free-text evidence into variables that are ready for analysis.

For human annotations in evaluation results, harmonization used a researcher-in-the-loop workflow. We sampled items for prompt calibration, compared prompt variants on the same sample, ran the selected prompt over the full set and shuffled subsets, and synthesized final dimensions and labels from recurring distinctions that were useful for analysis. Llama 3.3-70B was used for the harmonization mapping. Model-generated proposals were treated as evidence for researcher review rather than final labels. After final options were defined, open-ended items were mapped to harmonized labels, and ambiguous mappings were flagged for further inspection where applicable.

Some evaluation-results categories required specialized handling. Comparative results can be represented as tuples when the compared systems, comparison basis, direction, and evidence are clear. Performance-versus-human annotations are retained as metric-like source-category text rather than mapped into a qualitative label schema. Limitations are not further harmonized by default in the current release because the existing limitation categories already capture the intended analytic granularity. Machine annotations are harmonized source by source, using the same principle with a lighter workflow. For each model and category, harmonized options are generated from the annotations produced by that source and then used to map its open-ended items into comparable labels. 

Our hope is that the broad scientific community can use the benchmark experiments, released artifacts, and harmonized dataset together: to compare review-automation systems, audit model behavior, and study how the literature describes its domains, stages, approaches, results, and limitations.


\section*{Acknowledgments}

BL is supported by the U.S. National Institutes of Health (NIH), the U.S. Department of Veterans Affairs (VA), the Brain \& Behavior Research Foundation (BBRF), the Sidney R. Baer, Jr. Foundation, the Hasso Plattner Foundation, and the Windreich Family Foundation. No funding was specific to this work.

\bibliographystyle{unsrtnat}

\bibliography{main}  

\input{appendix}

\end{document}

%% file: appendix.tex
\clearpage
\appendix
\section*{Supplementary Information
}\setcounter{subsection}{0}\renewcommand{\thesubsection}{\Alph{subsection}}\renewcommand{\thesubsubsection}{\thesubsection.\arabic{subsubsection}}



\subsection{Search query}
\label{appendix:search_query}

\begin{searchquerybox}{Search query}
(
  TITLE-ABS-KEY(
    "literature review*" OR "systematic review*" 
    OR "scoping review*" OR "narrative review*" 
    OR "umbrella review*" OR "rapid review*" 
    OR "integrative review*" OR "evidence synthesis" 
    OR "meta-analysis"
  )
  AND TITLE-ABS-KEY(
    "large language model*" OR "LLM" OR "LLMs" OR "nlp" 
    OR "natural language processing" OR "transformer*"
    OR "AI" OR "artificial intelligence" OR "chatgpt" 
    OR "GPT-" OR "biogpt" OR "biomedgpt" OR "agrigpt" 
    OR "spiritualgpt" OR "lifegpt" OR "fingpt" OR "llama" 
    OR "llama-" OR "medllama" OR "mistral" OR "biomistral" 
    OR "mixtral" OR "mixtral-" OR "bard" OR "bard-" 
    OR "bert" OR "bert-" OR "legalbert" OR "rasonbert" 
    OR "finbert" OR "drbert" OR "biobert" OR "scibert" 
    OR "clinicalbert" OR "biomedbert" OR "mentalbert" 
    OR "pubmedbert" OR "claude" OR "claude-" OR "palm" 
    OR "palm-" OR "gemini" OR "gemini-" OR "gemma" 
    OR "gemma-" OR "copilot" OR "copilot-" OR "deepseek*" 
    OR "qwen" OR "qwen-" OR "phi" OR "phi-" OR "sciphi" 
    OR "scipphi" OR "falcon" OR "falcon-" OR "alpaca" 
    OR "alpaca-" OR "bloom" OR "bloom-" OR "grok" OR "grok-"
  )
) 
AND PUBYEAR < 2026 
AND NOT PUBDATETXT("July 2025" OR "August 2025")
\end{searchquerybox}

\subsection{Prompts}
\label{appendix:prompts}

\subsubsection{Title and abstract screening}

\label{appendix:prompts1}

\textbf{Basic:}
\begin{tcolorbox}[
    breakable,
    boxrule=0.5pt,
    sharp corners,
    fontupper=\small,
    colback=orange!5!white,
    colframe=orange!80!black,
    title={Basic System Prompt}
]
You are an expert in AI and literature reviews. \\
Given the TITLE and ABSTRACT of a scientific paper: 
Does this paper use AI to automate any part of the process of a
scientific review paper (of any kind)? \\
Answer with 'yes' or 'no' ONLY
\end{tcolorbox}

\textbf{Inclusion:}
\begin{tcolorbox}[
    breakable,
    boxrule=0.5pt,
    sharp corners,
    fontupper=\small,
    colback=orange!5!white,
    colframe=orange!80!black,
    title={Inclusion System Prompt}
]
You are an expert in AI and literature reviews. \\
Given the TITLE and ABSTRACT of a scientific paper: 
Does this paper use AI to automate any part of the process of a scientific review paper (of any kind)? \\
Include it only if the paper uses AI/ML/NLP/LLMs/LLMs to automate or algorithmically execute a step in evidence synthesis of scholarly articles (e.g., search, deduplication, title and abstract screening, full-text screening, data extraction, risk of bias, study classification, snowballing, meta-analysis). \\
Answer with 'yes' or 'no' ONLY
\end{tcolorbox}

\textbf{Inclusion + Exclusion:}
\begin{tcolorbox}[
    breakable,
    boxrule=0.5pt,
    sharp corners,
    fontupper=\small,
    colback=orange!5!white,
    colframe=orange!80!black,
    title={Inclusion + Exclusion System Prompt}
]
You are an expert in AI and literature reviews. \\
Given the TITLE and ABSTRACT of a scientific paper: 
Does this paper use AI to automate any part of the process of a scientific review paper (of any kind)? \\
Include it only if the paper uses AI/ML/NLP/LLMs/LLMs to automate or algorithmically execute a step in evidence synthesis of scholarly articles (e.g., search, deduplication, title and abstract screening, full-text screening, data extraction, risk of bias, study classification, snowballing, meta-analysis). \\
Exclude: domain literature reviews not about SR automation; literature review about the application of AI to a particular domain; bibliometric tools (e.g., citation-intent) unless directly used for SR steps. \\
Answer with 'yes' or 'no' ONLY
\end{tcolorbox}

\subsubsection{Full-text screening}
\label{appendix:prompts2}

\textbf{Unstructured:}
\begin{tcolorbox}[
    breakable,
    boxrule=0.5pt,
    sharp corners,
    fontupper=\small,
    colback=orange!5!white,
    colframe=orange!80!black,
    title={Unstructured System Prompt}
]
You are an expert in AI and literature reviews. \\
Given the full paper text, carefully analyze the content and discuss whether the paper uses AI to automate any part of a scientific review process.
\end{tcolorbox}

\textbf{Chain-of-Thought:}
\begin{tcolorbox}[
    breakable,
    boxrule=0.5pt,
    sharp corners,
    fontupper=\small,
    colback=orange!5!white,
    colframe=orange!80!black,
    title={Chain-of-Thought System Prompt}
]
You are an expert in AI and literature reviews. \\
Your task is to assess whether the paper uses AI to automate any part of a scientific review process. \\
1. First determine if the paper is conducting a literature review of any kind. \\
2. Second, assess if the paper uses AI or other NLP methods to automate some aspect of the review. \\
3. Finally, reach a conclusion of whether the paper is performing a scientific review using AI or not.
\end{tcolorbox}

\newpage

\subsection{Benchmark results}
\label{appendix:benchmark_results}

\input{table}\begin{table}[p]\centering\footnotesize\setlength{\tabcolsep}{3pt}\begin{tabularx}{\linewidth}{@{}>{\raggedright\arraybackslash}p{0.25\linewidth}>{\raggedright\arraybackslash}p{0.12\linewidth}*{4}{>{\centering\arraybackslash}X}@{}}\toprule\textbf{Model} & \textbf{Model Size} & \makecell{\textbf{Year}\\\textbf{Acc}} & \makecell{\textbf{Domain}\\\textbf{Jac}} & \makecell{\textbf{Review}\\\textbf{Stage Jac}} & \makecell{\textbf{Approach}\\\textbf{Jac}} \\\midrule Gemma 3 & 4B & 0.87 & 0.13 & 0.24 & 0.10 \\Gemma 3 & 12B & 0.91 & 0.29 & 0.37 & 0.19 \\Gemma 3 & 27B & 0.88 & 0.52 & 0.36 & 0.27 \\gpt-oss & 20B & 0.97 & 0.54 & 0.61 & 0.37 \\gpt-oss & 120B & 0.80 & 0.55 & 0.55 & 0.30 \\Llama 3 & 8B & 0.07 & 0.02 & 0.06 & 0.04 \\Llama 3 & 70B & 0.12 & 0.03 & 0.08 & 0.05 \\Llama 3 ChatQA & 70B & 0.00 & 0.05 & 0.01 & 0.00 \\Llama 3.1 & 8B & 0.89 & 0.20 & 0.30 & 0.15 \\Llama 3.1 & 70B & 0.96 & 0.63 & 0.57 & 0.27 \\Llama 3.2 & 1B & 0.33 & 0.06 & 0.13 & 0.08 \\Llama 3.2 & 3B & 0.57 & 0.16 & 0.18 & 0.09 \\Llama 3.3 & 70B & 0.97 & 0.65 & 0.51 & 0.23 \\Llama 4 & 16x17B & 0.15 & 0.05 & 0.03 & 0.04 \\Mistral Large & 123B & 0.67 & 0.36 & 0.39 & 0.26 \\Phi-3 & 3.8B & 0.49 & 0.10 & 0.19 & 0.06 \\Phi-3 & 14B & 0.81 & 0.32 & 0.36 & 0.07 \\Phi-3.5 & 3.8B & 0.60 & 0.17 & 0.24 & 0.11 \\Qwen3 & 8B & 0.91 & 0.60 & 0.55 & 0.27 \\Qwen3 & 14B & 0.93 & 0.57 & 0.58 & 0.30 \\Qwen3 & 30B & 0.97 & 0.70 & 0.57 & 0.36 \\Qwen3 & 32B & 0.92 & 0.64 & 0.55 & 0.36 \\\midrule\textbf{Best open-weight} & -- & 0.97 & 0.70 & 0.61 & 0.37 \\\bottomrule\end{tabularx}\caption{\textbf{Closed-field extraction performance of open-weight models on the full 888-paper set.} Year is scored by accuracy; domain, review stage, and approach are scored by Jaccard overlap. The ``Best open-weight'' row reports the highest value observed separately for each field and therefore does not represent a single deployable model.}\label{tab:de_closed}\end{table}\begin{table}[p]\centering\footnotesize\setlength{\tabcolsep}{3pt}\begin{tabularx}{\linewidth}{@{}>{\raggedright\arraybackslash}p{0.25\linewidth}>{\raggedright\arraybackslash}p{0.12\linewidth}*{4}{>{\centering\arraybackslash}X}@{}}\toprule\textbf{Model} & \textbf{Model Size} & \makecell{\textbf{Eval}\\\textbf{Recall}} & \makecell{\textbf{Eval}\\\textbf{Precision}} & \makecell{\textbf{Limit.}\\\textbf{Recall}} & \makecell{\textbf{Limit.}\\\textbf{Precision}} \\\midrule Gemma 3 & 4B & 0.161 & 0.454 & 0.056 & 0.074 \\Gemma 3 & 12B & 0.201 & 0.378 & 0.070 & 0.232 \\Gemma 3 & 27B & 0.196 & 0.584 & 0.009 & 0.046 \\gpt-oss & 20B & 0.251 & 0.785 & 0.120 & 0.229 \\gpt-oss & 120B & 0.230 & 0.808 & 0.131 & 0.200 \\Llama 3 & 8B & 0.003 & 0.092 & 0.005 & 0.048 \\Llama 3 & 70B & 0.012 & 0.915 & 0.009 & 0.095 \\Llama 3 ChatQA & 70B & 0.000 & -- & 0.000 & -- \\Llama 3.1 & 8B & 0.096 & 0.207 & 0.106 & -- \\Llama 3.1 & 70B & 0.191 & 0.511 & 0.208 & -- \\Llama 3.2 & 1B & 0.010 & 0.033 & 0.063 & 0.065 \\Llama 3.2 & 3B & 0.073 & 0.288 & 0.121 & 0.823 \\Llama 3.3 & 70B & 0.228 & 0.423 & 0.248 & 0.169 \\Llama 4 & 16x17B & 0.001 & 0.052 & 0.000 & -- \\Mistral Large & 123B & 0.207 & 0.772 & 0.020 & -- \\Phi-3 & 3.8B & 0.039 & 0.335 & 0.050 & 0.311 \\Phi-3 & 14B & 0.081 & 0.627 & 0.034 & 0.140 \\Phi-3.5 & 3.8B & 0.100 & 0.154 & 0.092 & 0.156 \\Qwen3 & 8B & 0.231 & 0.634 & 0.056 & 0.175 \\Qwen3 & 14B & 0.261 & 0.502 & 0.172 & 0.329 \\Qwen3 & 30B & 0.178 & 0.564 & 0.037 & -- \\Qwen3 & 32B & 0.300 & 0.448 & 0.159 & 0.208 \\\midrule\textbf{Best open-weight} & -- & 0.300 & 0.915 & 0.248 & 0.823 \\\bottomrule\end{tabularx}\caption{\textbf{Open-field recall and precision by model under calibrated LaaJ evaluation.} Dashes (--) indicate unsupported estimates per the criteria of \citet{laaj_calibration}; full criteria in Methods.}\label{tab:de_open}\end{table}

\clearpage

\subsection{Additional performance analysis}
\label{appendix:performance}

\begin{figure}[h]
\centering
\includegraphics[width=0.8\columnwidth]{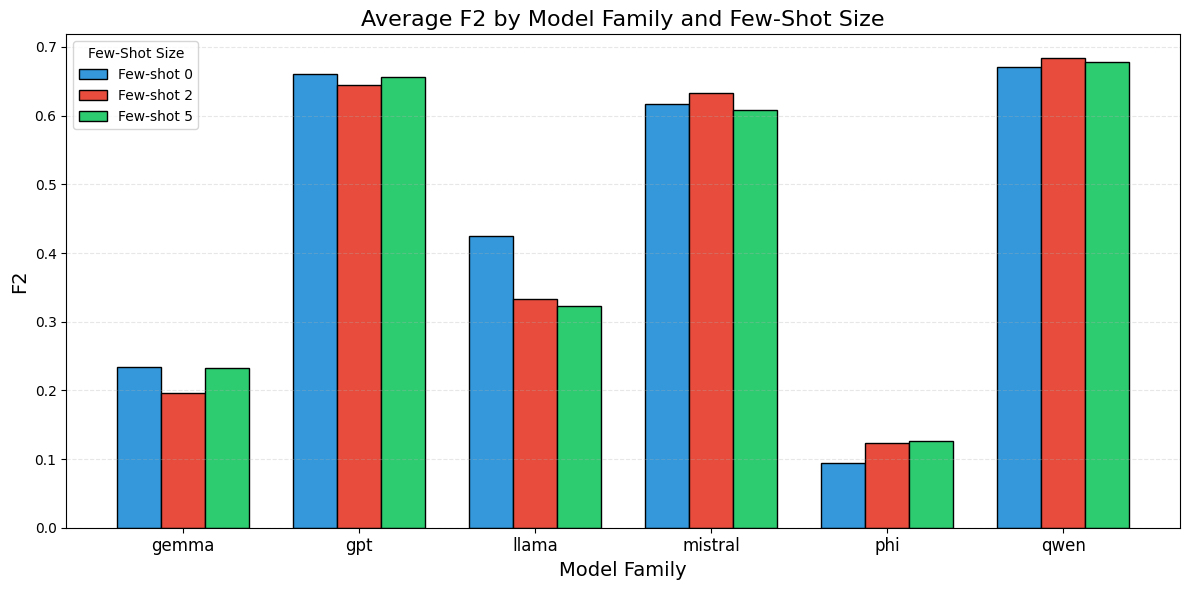}
\caption{\textbf{Effect of few-shot learning on title and abstract screening.} Average $F_2$ by model family when increasing the number of in-context examples (0, 2, 5). Gains are modest and heterogeneous across families: some improve slightly with a small number of examples, while others show no benefit or slight regressions, indicating few-shot prompting is not a universally reliable lever for TA screening.}
\label{fig:ta_fewshot}
\end{figure}

\begin{figure}[h]
\centering
\includegraphics[width=0.8\columnwidth]{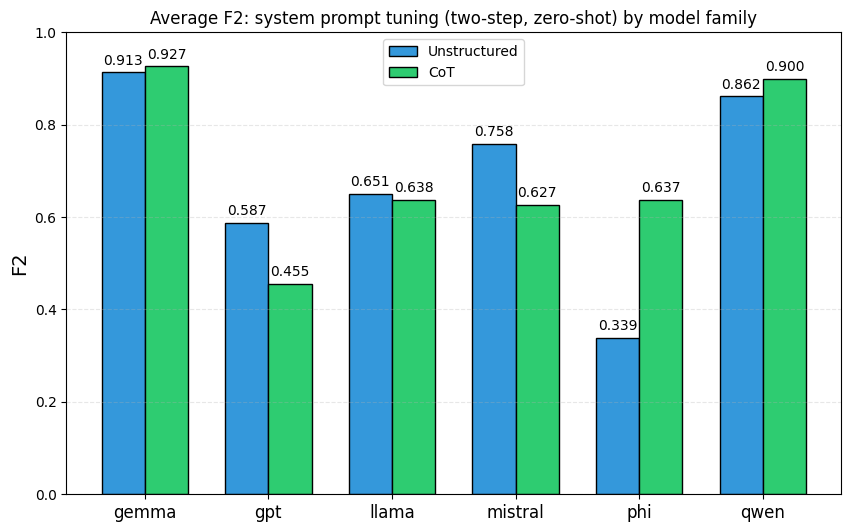}
\caption{\textbf{Effect of chain-of-thought (CoT) prompting in full-text screening.} Average $F_2$ by model family under two-step, zero-shot prompting, comparing an unstructured system prompt to a CoT variant that requires the model to reason through the criteria before emitting a binary decision. CoT yields a modest overall gain in $F_2$ (0.681 → 0.708), but effects are heterogeneous across families: it improves Gemma, Qwen, and Phi (notably lifting weaker families), while slightly degrading GPT, Llama, and Mistral on average.}
\label{fig:ft_cot}
\end{figure}


%% file: table.tex
\begin{table}[h]
\centering
\begin{adjustwidth}{-0.30in}{-0.30in}\centering

\scriptsize
\centering
\setlength{\tabcolsep}{10pt}
\renewcommand{\arraystretch}{1.05}

\begin{tabularx}{\linewidth}{@{} l c *{9}{C} @{}}
\toprule
& & \multicolumn{3}{c}{zero-shot}
  & \multicolumn{3}{c}{few-shot 2}
  & \multicolumn{3}{c}{few-shot 5} \\
\cmidrule(lr){3-5}\cmidrule(lr){6-8}\cmidrule(lr){9-11}
model & \makecell{model\\size}
& basic & incl & \makecell{incl+\\excl}
& basic & incl & \makecell{incl+\\excl}
& basic & incl & \makecell{incl+\\excl} \\
\midrule

Gemma 3 & 4B & 0.148 & 0.186 & 0.157 & 0.129 & 0.136 & 0.146 & 0.154 & 0.151 & 0.164 \\
Gemma 3 & 12B & 0.179 & 0.212 & 0.257 & 0.188 & 0.219 & 0.317 & 0.329 & 0.307 & 0.339 \\
Gemma 3 & 27B & 0.253 & 0.293 & 0.421 & 0.227 & 0.249 & 0.324 & 0.359 & 0.330 & 0.428 \\
gpt-oss & 20B & 0.620 & 0.655 & 0.628 & 0.644 & 0.658 & 0.679 & 0.677 & 0.694 & 0.686 \\
gpt-oss & 120B & 0.686 & 0.681 & 0.697 & 0.687 & 0.684 & 0.675 & 0.674 & 0.659 & 0.686 \\
Llama 3 & 8B & 0.178 & 0.360 & 0.452 & 0.205 & 0.312 & 0.476 & 0.310 & 0.377 & 0.542 \\
Llama 3 & 70B & 0.438 & 0.567 & 0.679 & 0.283 & 0.358 & 0.570 & 0.288 & 0.471 & 0.698 \\
Llama 3 ChatQA & 70B & 0.268 & 0.213 & 0.287 & 0.322 & 0.390 & 0.472 & 0.315 & 0.410 & 0.466 \\
Llama 3.1 & 8B & 0.295 & 0.535 & 0.609 & 0.210 & 0.456 & 0.646 & 0.459 & 0.534 & 0.588 \\
Llama 3.1 & 70B & 0.421 & 0.598 & 0.755 & 0.395 & 0.577 & 0.751$^+$ & 0.302 & 0.531 & 0.749$^+$$^+$ \\
Llama 3.2 & 1B & 0.115 & 0.195 & 0.186 & 0.174 & 0.181 & 0.197 & 0.128 & 0.132 & 0.148 \\
Llama 3.2 & 3B & 0.327 & 0.579 & 0.564 & 0.292 & 0.335 & 0.410 & 0.318 & 0.299 & 0.430 \\
Llama 3.3 & 70B & 0.251 & 0.497 & 0.786$^*$ & 0.238 & 0.384 & 0.717$^+$$^+$$^+$ & 0.228 & 0.464 & 0.735$^+$ \\
Llama 4 & 16x17B & 0.345 & 0.478 & 0.482 & 0.431 & 0.620 & 0.737 & 0.556 & 0.625 & 0.669 \\
Mistral Large & 123B & 0.637 & 0.610 & 0.606 & 0.766$^*$ & 0.671 & 0.570 & 0.728 & 0.638 & 0.614 \\
Phi-3 & 3.8B & 0.023 & 0.076 & 0.077 & 0.147 & 0.164 & 0.245 & 0.192 & 0.217 & 0.258 \\
Phi-3 & 14B & 0.206 & 0.217 & 0.246 & 0.179 & 0.355 & 0.477 & 0.207 & 0.271 & 0.295 \\
Phi-3.5 & 3.8B & 0.000 & 0.000 & 0.000 & 0.000 & 0.000 & 0.000 & 0.065 & 0.043 & 0.041 \\
Qwen3 & 8B & 0.666 & 0.646 & 0.665 & 0.686 & 0.660 & 0.667 & 0.638 & 0.665 & 0.725 \\
Qwen3 & 14B & 0.761$^*$ & 0.720 & 0.749 & 0.739$^+$ & 0.753 & 0.736 & 0.749 & 0.756$^*$ & 0.736 \\
Qwen3 & 30B & 0.706 & 0.567 & 0.565 & 0.726$^+$ & 0.760$^*$ & 0.712 & 0.734 & 0.734 & 0.751 \\
Qwen3 & 32B & 0.657 & 0.682 & 0.668 & 0.754 & 0.704 & 0.728 & 0.706 & 0.726 & 0.740 \\

\midrule
\multicolumn{2}{l}{strong pool ($t_\text{s}$=3)} & \multicolumn{9}{c} {\textbf{0.808} (precision: \textbf{0.551}, recall: 0.915)}  \\
\multicolumn{2}{l}{lenient pool ($t_\text{l}$=9)} & \multicolumn{9}{c} {0.803 (precision: 0.512, recall: 0.936)}  \\
\multicolumn{2}{l}{comb. pool ($t_{\text{s}}=3$, $t_{\text{l}}=4$)} & \multicolumn{9}{c} {0.655 (precision: 0.275, recall: \textbf{1.000})}  \\
\bottomrule

\end{tabularx}
\caption{\textbf{Systematic evaluation in title and abstract screening.} We report $F_2$ (recall-weighted $F$-score) for each model across prompting and few-shot configurations. From single-model results, we select the top five configurations to form a \emph{strong pool} (marked $^*$) and build a majority-vote ensemble (predict “include” if at least $t_\text{s}$ of 5 vote yes). To prioritize high recall, we also form a \emph{lenient pool} (marked $^+$) of ten configurations with the highest recall among those with precision above 0.2, excluding configurations already in the strong pool; pool membership is frozen to resolve ties reproducibly. We further evaluate a \emph{combined} ensemble that predicts “include” if either pool crosses its threshold. For these three ensemble variants we report $F_2$, precision, and recall, and select the best system subject to recall $=1.0$ for curation.}
\label{tab:stage_A}
\vskip -0.1in
\end{adjustwidth}
\end{table}

\begin{table}[h]
\centering
\begin{adjustwidth}{-0.30in}{-0.30in}\centering
\scriptsize
\setlength{\tabcolsep}{8pt}
\begin{tabularx}{\linewidth}{>{\raggedright\arraybackslash}l X *{8}{l}}
\toprule
& & \multicolumn{4}{c}{zero-shot} & \multicolumn{4}{c}{few-shot} \\
\cmidrule(lr){3-6}\cmidrule(lr){7-10}
model & model size & \multicolumn{2}{c}{one-step} & \multicolumn{2}{c}{two-step} & \multicolumn{2}{c}{no thought traces} & \multicolumn{2}{c}{thought traces} \\
\cmidrule(lr){3-4}\cmidrule(lr){5-6}\cmidrule(lr){7-8}\cmidrule(lr){9-10}
& & basic & strict & unstruct. & CoT & few-shot 2 & few-shot 4 & few-shot 2 & few-shot 4 \\
\midrule
Gemma 3 & 4B & 0.661 & 0.877 & 0.927 & 0.921 & 0.702 & 0.761 & 0.854 & 0.809 \\
Gemma 3 & 12B & 0.434 & 0.605 & 0.893 & 0.911 & 0.793 & 0.624 & 0.855 & 0.829 \\
Gemma 3 & 27B & 0.908 & 0.808 & 0.920 & 0.948 & 0.817 & 0.691 & 0.865 & 0.761 \\
gpt-oss & 20B & 0.908 & 0.882 & 0.897 & 0.901 & 0.865 & 0.784 & 0.910 & 0.941 \\
gpt-oss & 120B & 0.649 & 0.644 & 0.277 & 0.009 & 0.286 & 0.127 & 0.254 & 0.228 \\
Llama 3 & 8B & 0.126 & 0.134 & 0.630 & 0.628 & 0.673 & 0.673 & 0.673 & 0.673 \\
Llama 3 & 70B & 0.143 & 0.143 & 0.776 & 0.769 & 0.770 & 0.770 & 0.746 & 0.746 \\
Llama 3 ChatQA & 70B & 0.000 & 0.000 & 0.056 & 0.046 & 0.046 & 0.046 & 0.073 & 0.082 \\
Llama 3.1 & 8B & 0.928 & 0.913 & 0.932 & 0.916 & 0.900 & 0.027 & 0.856 & 0.940 \\
Llama 3.1 & 70B & 0.976 & 0.975 & 0.972 & 0.970 & 0.689 & 0.108 & 0.934 & 0.951 \\
Llama 3.2 & 1B & 0.027 & 0.450 & 0.181 & 0.301 & 0.217 & 0.047 & 0.351 & 0.793 \\
Llama 3.2 & 3B & 0.785 & 0.780 & 0.897 & 0.772 & 0.852 & 0.975 & 0.946 & 0.933 \\
Llama 3.3 & 70B & 0.976 & 0.975 & \textbf{0.981} & 0.975 & 0.955 & 0.923 & 0.936 & 0.957 \\
Llama 4 & 16x17B & 0.009 & 0.018 & 0.431 & 0.362 & 0.549 & 0.461 & 0.503 & 0.496 \\
Mistral Large & 123B & 0.696 & 0.697 & 0.758 & 0.627 & 0.776 & 0.781 & 0.775 & 0.769 \\
Phi-3 & 3.8B & 0.329 & 0.065 & 0.164 & 0.391 & 0.970 & 0.968 & 0.845 & 0.791 \\
Phi-3 & 14B & 0.287 & 0.600 & 0.644 & 0.734 & 0.968 & 0.956 & 0.956 & 0.931 \\
Phi-3.5 & 3.8B & 0.229 & 0.543 & 0.209 & 0.787 & 0.515 & 0.917 & 0.861 & 0.859 \\
Qwen3 & 8B & 0.882 & 0.917 & 0.896 & 0.937 & 0.901 & 0.884 & 0.878 & 0.912 \\
Qwen3 & 14B & 0.929 & 0.933 & 0.888 & 0.902 & 0.889 & 0.902 & 0.888 & 0.923 \\
Qwen3 & 30B & 0.727 & 0.744 & 0.737 & 0.823 & 0.866 & 0.734 & 0.798 & 0.823 \\
Qwen3 & 32B & 0.901 & 0.899 & 0.926 & 0.938 & 0.810 & 0.874 & 0.864 & 0.919 \\
\bottomrule
\end{tabularx}
\caption{\textbf{Systematic evaluation in full-text screening.} We report $F_2$ (the recall-weighted F-score) for each model across prompting and few-shot configurations. We also evaluate ensembling variants—including majority-vote ensembles over a \emph{strong pool} of the top five single-model configurations and a \emph{lenient pool} of configurations achieving recall (=1.0), a \emph{combined} ensemble that triggers inclusion if either pool meets its threshold, and a token-length–aware \emph{routing} ensemble (logistic regression with context-length bins). In our setting, none of these ensembles improves over the best single model, Llama~3.3-70B in zero-shot with the unstructured system prompt; we therefore use this configuration to score the remaining unlabeled full texts and manually re-annotate predicted positives for downstream extraction corpus curation.}
\label{tab:stage_B}
\vskip -0.1in
\end{adjustwidth}
\end{table}
